\documentclass{article}
\usepackage{hyperref}
\usepackage{graphicx}

\title{Towards Democratizing AI: A Comparative Analysis of AI-as-a-Service Platforms and the Open Space for Machine Learning Approach}
\author{Dennis Rall\textsuperscript{1}, Bernhard Bauer\textsuperscript{2}, Thomas Fraunholz\textsuperscript{1}}
\date{\today}

\begin{document}

\maketitle

\begin{abstract}
Recent AI research has significantly reduced the barriers to apply AI, but the process of setting up the necessary tools and frameworks can still be a challenge. While AI-as-a-Service platforms have emerged to simplify the training and deployment of AI models, they still fall short of achieving true democratization of AI. In this paper, we aim to address this gap by comparing several popular AI-as-a-Service platforms and identifying the key requirements for a platform that can achieve true democratization of AI. Our analysis highlights the need for self-hosting options, high scalability, and openness. To address these requirements, we propose our approach: the "Open Space for Machine Learning" platform. Our platform is built on cutting-edge technologies such as Kubernetes, Kubeflow Pipelines, and Ludwig, enabling us to overcome the challenges of democratizing AI. We argue that our approach is more comprehensive and effective in meeting the requirements of democratizing AI than existing AI-as-a-Service platforms.
\end{abstract}


\footnotetext[1]{WOGRA AG, Hery-Park 3000, 86368 Gersthofen, Germany}
\footnotetext[2]{University of Augsburg, Germany}

\section{Introduction}\label{introduction}

The democratization of AI is currently a trend in AI research, driven by
the shortage of AI experts that hinders the use of AI in many areas,
including stock marking trading {[}1{]} and personalized medicine
{[}2{]}, to name only a couple. The aim is to reduce the required
expertise and make AI more accessible to a wider range of users.
Significant progress has been made in the development and training of AI
models, enabling the creation and training of models automatically from
data, known as low-code AI. Examples of low-code tools include Ludwig
{[}3{]}, fast-ai {[}4{]}, and Autogluon {[}5, 6{]}. Low-code AI tools
can be coupled with user-friendly interfaces to transition to no-code
AI. These advancements hold the potential to make AI accessible to
people with little or no programming experience, allowing them to apply
AI to various tasks without the need for technical expertise.

However, providing the infrastructure and setting up these frameworks
and tools remains a hurdle that requires expert knowledge. To overcome
this obstacle and further promote the democratization of AI, platforms
are emerging that take care of these tasks for the user and offer AI as
a service. These platforms provide a range of services, such as data
storage, model training, and API access, allowing users to easily
integrate AI into their workflow without the need for significant
technical expertise. With AI-as-a-Service, individuals and organizations
can benefit from the power of AI without the need to invest in expensive
hardware, software, or personnel. As a result, AI-as-a-Service has the
potential to accelerate the adoption of AI and make its benefits more
widely available.

In this article, we aim to compare existing AI-as-a-Service platforms
and examine why they have not yet led to a breakthrough in the
democratization of AI. Based on this analysis, we aim to identify the
requirements for a platform that can successfully achieve this goal and
then discuss our implementation to meet these requirements. By exploring
the strengths and weaknesses of existing AI-as-a-Service platforms, we
hope to provide insights into the design and development of a platform
that can truly democratize AI and make its benefits accessible to
everyone. Through this analysis, we hope to contribute to the ongoing
efforts to democratize AI and make its benefits available to a wider
audience.

\section{Study of Existing
Platforms}\label{study-of-existing-platforms}

The availability of easy-to-use AutoML solutions as a platform is a
crucial factor for the participation of non-AI experts in the technology
of machine learning. We will now take a look at the various existing
platforms for AutoML.

\subsection{AutoML Platforms by major Cloud
Providers}\label{automl-platforms-by-major-cloud-providers}
The major cloud providers - Amazon SageMaker\footnote{\url{https://aws.amazon.com/sagemaker/}}, Google Vertex AI\footnote{\url{https://cloud.google.com/vertex-ai}}, and Microsoft Azure Automated ML\footnote{\url{https://azure.microsoft.com/en-us/products/machine-learning/automatedml/}} - have invested significantly in AutoML and have developed powerful and scalable solutions. These machine learning platforms leverage cutting-edge algorithms and state-of-the-art technologies. Additionally, these platforms offer a high level of service, from prototyping to production, and provide a no-code environment that enables developers to easily build and deploy machine learning models.

One major disadvantage of these platform solutions is their lack of
openness. Since these platforms are not open source, expert users are
limited in their ability to customize and modify algorithms to fit their
specific needs.

Additionally, these platforms are not self-hosted, meaning that
sensitive data must be entrusted to a third-party provider.

Furthermore, the algorithms used by these platforms are often black
boxes, making it difficult to understand how they arrived at their
conclusions. This lack of transparency can be problematic, particularly
when it comes to ethical considerations and ensuring that the algorithms
are not biased or discriminatory.

Additionally, customers are often locked into using a single vendor,
limiting their ability to switch providers or take their business
elsewhere. Finally, while these platforms offer a no-code environment,
they are often complex and difficult to navigate without a solid
understanding of machine learning concepts and terminology. This can be
a major barrier for non-experts trying to leverage machine learning in
their work.

\subsection{Standalone AutoML
Platforms}\label{standalone-automl-platforms}

An established platform that has made it its mission to overcome this
obstacle is DataRobot\footnote{\url{https://www.datarobot.com/}}. The
platform stands out in terms of user-friendliness. Its no-code
environment makes it easy for non-experts to build and deploy machine
learning models, and its scalable AutoML solution covers the entire
machine learning life cycle. However, like the major cloud providers,
DataRobot's platform is not open source, meaning that experts are
limited in their ability to customize algorithms. Additionally, the
platform has no option to self-host, and the algorithms used are often
black boxes, limiting transparency and posing ethical concerns. Finally,
users are locked into the DataRobot ecosystem, which can be a major
drawback.

On the other hand, Predibase\footnote{\url{https://predibase.com/}}
attempts to achieve complete transparency using a declarative machine
learning approach {[}7{]} and utilizing open-source components under the
hood. However, minimal programming knowledge is required for the
low-code platform. In addition to this, is not foreseeable that the
platform will be made available as open source. Thus, it cannot be
self-hosted for sensitive data.

For users looking for a more open approach, H2O\footnote{\url{https://h2o.ai/}}
{[}8{]} is a powerful option. As an AutoML platform, H2O provides a
scalable solution for the complete machine learning lifecycle that can
be used either as a PaaS or self-hosted. However, H2O's platform does
have some limitations. Its low-code environment may not be accessible to
users with no programming skills, limiting the democratization of
machine learning. H2O knows about this drawback and introduced a
commercial no-code offering, H2O Driverless AI Wizard, which leverages
the full potential of AutoML for users with no programming. But this
solution is not open-source and cannot be self-hosted.

\section{Requirements}\label{requirements}

So far, we have gained insights into the various platforms that offer AI
as a service, and we have concluded that they are not yet sufficient for
achieving true AI democratization. Therefore, our next step is to derive
the requirements for such a platform that can lead us towards the
democratization of AI. These requirements will be crucial for creating a
platform that can enable people from diverse backgrounds to access and
benefit from AI technologies. We will focus on identifying the key
features and functionalities that are needed to build a platform that
can support a wide range of users, including those with limited
technical expertise.

\subsection{Platform}\label{platform}

But first, we want to emphasize why such a platform that offers AI as a
service is a requirement for achieving true AI democratization. One of
the main reasons why such a platform is crucial is that it allows users
to start using the software without the need for a complex setup
process. Especially with AI applications, there may be a significant
amount of expert knowledge required, as certain parts of the application
need to be executed on specialized hardware, such as GPUs, for effective
performance. By offering AI as a service, users can access the power of
AI without needing to invest in expensive hardware or hire specialized
personnel to configure and operate the system.

\subsection{Easy-to-use
User-Interface}\label{easy-to-use-user-interface}

In order to enable users with limited technical expertise to use the
platform, it is important to ensure that the user experience is
optimized. The platform should have an intuitive user interface that is
easy to navigate. With just a few interactions, users should be able to
train an AI model based on their own data without needing any expert
knowledge of AI or complex technical terms. By abstracting away the
complexities of AI, the platform can become more accessible to a wider
range of users, regardless of their level of technical expertise.

\subsection{Scalability}\label{scalability}

As there are many areas where the potential of AI has not yet been fully
exploited, a platform offering AI as a service could have a vast user
base, making scalability essential. With the increasing amount of data
being generated and stored, the platform must also be capable of
handling large amounts of data. Additionally, the complexity of AI
models is increasing, which requires more computing power for training,
so the platform must be able to handle increasingly complex AI
algorithms. In summary, the platform should be scalable in terms of its
user base, data handling capacity, and ability to handle complex AI
models.

\subsection{Adaptability and
Extensibility}\label{adaptability-and-extensibility}

Beyond the requirement for scalability in handling more complex AI
algorithms, there is another important aspect to consider: the
platform's adaptability and extensibility. This is crucial because the
field of AI is constantly evolving, with ongoing research and new
advancements being made. If the platform is designed in a way that
enables easy integration of these new developments, it can benefit a
wider range of users. By allowing for customization and flexibility in
the development of new AI models and algorithms, the platform can
continue to evolve alongside the latest advancements in AI, ensuring
that users have access to the most up-to-date and relevant tools and
technologies. It is important to note that this adaptability and
extensibility need not necessarily be achieved through the user
interface alone. As AI experts are required to implement and integrate
these changes, they can be made directly at the code level.

\subsection{Openness}\label{openness}

Having open-source code is therefore a requirement for the platform so
that these changes can be made by as many AI experts as possible.
Generally, public access to the platform's code offers further
advantages. Firstly, it increases trust in the platform, as independent
experts can verify the internal workings of the platform. Secondly, the
platform also benefits from an open-source community around it. This
way, it can be developed faster, more efficiently, and cost-effectively,
and users can help each other with questions and problems.

\subsection{Self-Hosting}\label{self-hosting}

Furthermore, in addition to the publicly available instance of the
platform, it may be important for some users to be able to deploy their
own instance. This allows for greater security considerations, as access
can be heavily restricted. So, the platform can also be used for data
that requires a higher level of security, expanding the use of AI in
such areas. This is also another argument for making the code of the
platform open-source, as it enables the tracking of what precisely
occurs with the uploaded data.

\section{Technologies}\label{technologies}

Now that we have identified the necessary requirements for a platform
that offers AI as a service, we will take a look at the technologies
that we will use to implement our own platform.

\subsection{Kubernetes}\label{kubernetes}

In order to address the significant scaling requirements of our
platform, we will be utilizing the open-source container orchestration
framework Kubernetes {[}9{]}. By deploying Kubernetes in the cloud, we
can take advantage of the auto-scaling mechanisms provided by cloud
providers.

Kubernetes provides a high level of abstraction by organizing available
computers as nodes and grouping them into a Kubernetes cluster. The
platform's Control Plane API enables users to manage the state of their
containers, while the Control Plane itself distributes the workload in
the form of pods to the nodes. Pods are the smallest unit in Kubernetes
and consist of one or more containers that run on the nodes {[}10{]}.
This approach allows us to efficiently manage the scaling of our
platform and ensure optimal resource utilization.

\subsection{Service Mesh}\label{service-mesh}

A service mesh is a software infrastructure that enables managing and
operating microservices-based applications. It provides a range of
benefits, such as improved scalability, reliability, and security. By
using a service mesh, application developers and operators can reduce
the complexity of their architecture while improving flexibility and
agility. A service mesh allows developers to focus on application
development without having to worry about the underlying infrastructure,
while providing operators with better control over application traffic
and security.

One of the main advantages of using a service mesh is its ability to
provide advanced traffic management capabilities. With a service mesh,
operators can manage and control the flow of traffic between services,
allowing them to balance the load and ensure optimal performance.
Additionally, a service mesh can also provide improved security features
such as authentication and authorization, allowing only authorized
traffic to pass through.

Istio {[}11{]} is a popular open-source service mesh that provides a
variety of features such as load balancing, traffic management, security
features like authentication and authorization, as well as
troubleshooting and monitoring. By using Istio, developers and operators
can reduce the complexity of microservices-based applications by
managing traffic between services within the application. Overall, Istio
is a powerful and flexible service mesh that helps developers and
operators operate microservices-based applications more effectively and
securely.

\subsection{Kubeflow Pipelines}\label{kubeflow-pipelines}

While simple microservices can be deployed as native Kubernetes
deployments, our machine learning tasks require special treatment due to
their need for significant computing power and specialized hardware. To
address this, we will be using a workflow engine, specifically Kubeflow
Pipelines {[}12{]}, which is built on top of Argo Workflow {[}13{]} - a
workflow engine designed specifically for Kubernetes environments. With
Argo Workflow, users can easily create, manage, and run complex
workflows, including creating workflow templates and executing them with
different parameters.

Kubeflow Pipelines provides additional ML functionality such as an
artifact store for data and models, visualizations for executions and
metrics, and robust features that increase the reliability of our
workflows, such as automatic retries and error handling. By using
Kubeflow, we can ensure that our ML workflows can handle errors and
unexpected events gracefully and achieve greater efficiency and
reliability in our development process.

Kubeflow Pipelines is a component of the Kubeflow project, which is
designed to facilitate the development, deployment, and management of
machine learning models on Kubernetes. In addition to the pipelines
component, it includes support for Jupyter Notebooks to explore data and
an interface to TensorFlow to train and serve models. Kubeflow also
offers model versioning and monitoring, enabling users to keep track of
their models' performance over time.

\subsection{Ludwig}\label{ludwig}

For our No-Code AI component, we will make use of Ludwig {[}3{]}, a
declarative ML framework {[}7{]} that creates and trains ML models using
a configuration file as input. This configuration file specifies the
input and target data types, as well as other parameters needed for
training, such as the optimizer, the loss function, and the number of
epochs. Ludwig's declarative approach eliminates the need to write
low-level code, making it ideal for non-technical users who want to
build and train ML models easily.

This can best be demonstrated through an example, using the PetFinder
dataset {[}14{]}. This dataset contains information about animals in a
shelter, such as their name, age, or an image of the animal. The goal is
to predict how quickly these animals will be adopted, which is
represented as categorical values in the dataset. An AdoptionSpeed of 4,
for example, indicates that the animal has not been adopted within 100
days. An example is shown in the following figure~\ref{fig:pet}, and 
a Ludwig configuration for this task is shown in the next figure~\ref
{fig:ludwig}. The configuration file defines the input and output
columns and sets some additional training parameters.

\begin{figure}
  \centering
  \includegraphics[width=\textwidth]{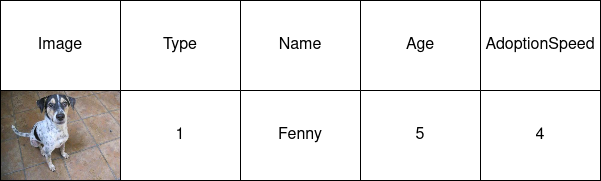}
  \caption{Example of a pet from the PetFinder dataset {[}14{]}. Fenny is a 5-year-old dog (Type 1) that has not been adopted within 100 days
  (AdoptionSpeed 4)}
  \label{fig:pet}
\end{figure}

\begin{figure}
  \centering
  \includegraphics[width=0.25\textwidth]{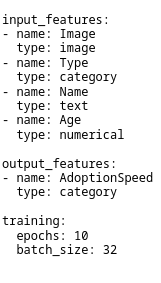}
  \caption{An example Ludwig configuration for the PetFinder dataset
  {[}14{]}. The input and output features are listed, along with some
  additional training information.}
  \label{fig:ludwig}
\end{figure}

By providing intuitive selection options in a user interface, we
eliminate the need for manual writing of the configuration file.
Additionally, we remove additional training parameters from the
configuration process in favor of employing techniques of hyperparameter
optimization. This approach expands the search space for AI models,
reducing the potential for bias caused by an expert's experience. Using
a user interface, we can thus expand this low-code ML tool into a
no-code tool. This enables us to allow users with no technical
experience to interact with our platform.

\section{Architecture}\label{architecture}

\begin{figure}
  \centering
  \includegraphics[width=0.55\textwidth]{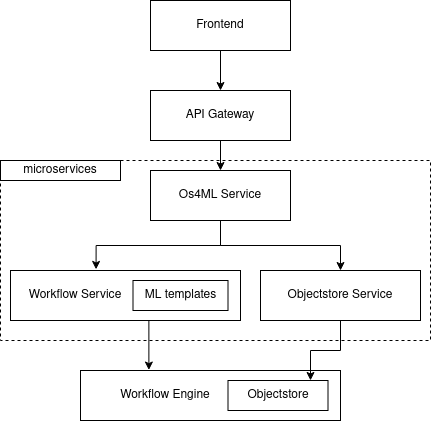}

  \caption{Our architecture centers around the API Gateway, which serves as the backbone of our platform by securely granting access to our microservices, including the Os4ML Service, the Workflow Service, and the Objectstore Service.}
  \label{fig:architecture}
\end{figure}

We will now explain how we combine various technologies to create our
AI-as-a-Service platform called Open Space for Machine Learning (Os4ML).
Figure~\ref{fig:architecture} illustrates the architecture of our platform.

Our platform is designed using microservices as its architectural
pattern. We have structured these microservices into a mesh topology
that is accessible through an API gateway, which we have implemented
using Istio. This architecture provides a flexible and scalable platform
that enables developers to access different microservices conveniently.
The frontend and other components can easily access the microservices
through the API gateway, which simplifies development and maintenance
tasks.

One of the most crucial microservices in our platform is the
Os4ml-Service. It is responsible for managing our domain models and
storing critical information such as metrics and execution times. This
service plays a vital role in providing efficient and reliable machine
learning solutions.

In addition, we have implemented the objectstore service, which offers a
dependable way to store and retrieve models and data. Furthermore, we
have the workflow service, which handles communication with the workflow
engine. For our platform, we use Kubeflow Pipelines as our workflow
engine. Both the workflow engine and the objectstore service can share
the same objectstore, which ensures consistency and reliability in our
platform.

Our frontend allows the user to upload data. In the background, the data
is transformed into an interface that is compatible with Ludwig (e. g.,
into Pandas dataframes), with data types detected and saved. We provide
templates for the workflow engine to create a Ludwig configuration and
train a machine learning model using the dataframe. These templates can
be quickly and easily customized to create tailored machine learning
models. Our platform thus provides a user-friendly and effective way to
train and deploy machine learning models.

To ensure scalability, we deploy the entire system on Kubernetes, which
provides a robust and reliable platform for container orchestration.
With Kubernetes, our system can handle large volumes of data and easily
adapt to changes in demand. By breaking down the system into smaller,
independent components, we can maintain better control and reduce the
risk of system-wide failures. This architecture enables a flexible and
scalable system that can efficiently handle complex machine learning
workloads.

\section{Deployment}\label{deployment}

The requirements section has highlighted the need to allow the advanced
users to deploy the platform themselves. To achieve this, we have
adopted a three-part Infrastructure-as-Code approach. In general,
Infrastructure-as-Code involves provisioning the necessary
infrastructure using code so that it can be reproduced or restored in a
predictable manner. To implement this, we use the open-source tool
Terraform {[}15{]}. Our deployment process is divided into three steps:
cloud setup, Kubernetes cluster setup, and the actual deployment of our
platform. If steps 1 and/or 2 are not required because they already
exist, the user can skip directly to step 3.

We will be publishing the Infrastructure-as-Code scripts along with our
application code. This means that deploying the platform will only
require a single command, in our case, 'terraform apply'. This will
greatly reduce the time and effort required for the deployment process,
making it more efficient and user-friendly. Additionally, this approach
ensures consistency in the infrastructure setup, reducing the
possibility of errors caused by manual configuration.

In order to realize step 3 of our deployment process, we utilize the
declarative GitOps continuous delivery tool ArgoCD {[}16{]}. This
provides the administrators of individual instances with access to
updates to the platform as well as a pre-packaged monitoring solution.
By utilizing ArgoCD, we can automate the deployment process, allowing
for seamless updates and streamlined management of our platform.
Additionally, the use of declarative configurations ensures that our
infrastructure remains consistent across all environments, reducing the
risk of configuration drift and ensuring high reliability.

\hypertarget{accomplishments-and-contributions}{%
\section{Accomplishments and
Contributions}\label{accomplishments-and-contributions}}

We are thrilled to report significant progress made on our
AI-as-a-Service platform, "Open Space for Machine Learning"\footnote{\url{https://www.os4ml.com/}},
and are eager to share the goals we've achieved and contributions we've
made.

\subsection{Transforming Ludwig from Low-Code to
No-Code}\label{transforming-ludwig-from-low-code-to-no-code}

Our most noteworthy accomplishment is the creation of a user-friendly
frontend that has transformed Ludwig, our low-code ML tool, into a
no-code tool. With our efforts, we have opened up the platform to a
broader range of users, including those with little to no experience in
machine learning, who can now effortlessly train and deploy AI models
without the need to write any code. Our innovation has revolutionized
the industry, making machine learning accessible to anyone.

\subsection{Integrating Multiple Open-Source
Tools}\label{integrating-multiple-open-source-tools}

Another proud achievement is our integration of several powerful
open-source tools, such as Istio, Kubernetes, and Kubeflow, into our
platform, creating a blueprint for highly scalable AI applications. This
has opened up new avenues for exploring the potential of AI.

\subsection{Cloud-Enabled Platform with Self-Hosting
Option}\label{cloud-enabled-platform-with-self-hosting-option}

Furthermore, we have made our platform cloud-agnostic and easy to deploy
for the community. This approach allows us to take advantage of the
benefits of cloud computing, such as scalability and cost-effectiveness,
while still providing the option for users to self-host our platform
on-premises. By enabling easy deployment, we have made it simpler for
the community to take advantage of our AI-as-a-Service platform,
including those dealing with sensitive data.

Overall, we are proud of our contributions to the AI community and look
forward to continuing to push the boundaries of what is possible in the
world of machine learning.

\section{Summary \& Outlook}\label{summary-outlook}

In this article, we have demonstrated that the next step in AI
democratization is the provision of AI as a service through a platform.
However, current solutions have not been satisfactory, leading us to
collect requirements for such a platform. We have shown how we use
existing open-source solutions and add our own components to meet the
collected requirements. Our platform is also available as open
source\footnote{\url{https://github.com/WOGRA-AG/Os4ML}}, enabling
further collaboration and development towards more accessible AI
solutions. Now that the foundational work is complete, the next step is
to enhance the AI component. One area of improvement would be to
incorporate an explainable AI mechanism. This would provide valuable
insights into the trained model's decision-making process, thereby
increasing trust and transparency.

\section*{Acknowledgments}
Os4ML is a project of the WOGRA AG research group in cooperation with
the German Aerospace Center and is funded by the Ministry of Economic
Affairs, Regional Development, and Energy as part of the High Tech
Agenda of the Free State of Bavaria.

\section*{References}
\begin{enumerate}
  \def\labelenumi{\arabic{enumi}.}
  \item
    Jan Juricek. 2014. The Use of Artificial Intelligence in Building
    Automated Trading Systems. \emph{International Journal of Computer
    Theory and Engineering} 6, (January 2014), 326--329.DOI: \url{https://doi.org/10.7763/IJCTE.2014.V6.883}
  \item
    Jamilu Awwalu, Ali Garba, Anahita Ghazvini, and Rose Atuah. 2015.
    Artificial Intelligence in Personalized Medicine Application of AI
    Algorithms in Solving Personalized Medicine Problems.
    \emph{International Journal of Computer Theory and Engineering} 7,
    (December 2015), 439--443. DOI:\url{https://doi.org/10.7763/IJCTE.2015.V7.999}
  \item
    Piero Molino, Yaroslav Dudin, and Sai Sumanth Miryala. 2019.
    \emph{Ludwig: a type-based declarative deep learning toolbox}.
  \item
    Jeremy Howard and Sylvain Gugger. 2020. Fastai: A Layered API for Deep
    Learning. \emph{Information} 11, 2 (2020). DOI:\url{https://doi.org/10.3390/info11020108}
  \item
    Nick Erickson, Jonas Mueller, Alexander Shirkov, Hang Zhang, Pedro
    Larroy, Mu Li, and Alexander Smola. 2020. \emph{AutoGluon-Tabular:
    Robust and Accurate AutoML for Structured Data.}
  \item
    Xingjian Shi, Jonas Mueller, Nick Erickson, Mu Li, and Alexander J.
    Smola. 2021. \emph{Benchmarking Multimodal AutoML for Tabular Data
    with Text Fields}.
  \item
    Piero Molino and Christopher Ré. 2021. \emph{Declarative Machine
    Learning Systems}.
  \item
    H2O.ai. Retrieved March 29, 2023 from
    https://github.com/h2oai/h2o-3
  \item
    Kubernetes Components. Retrieved March 29, 2023 from \url{https://kubernetes.io/docs/concepts/overview/components/}
  \item
    Shazibul Islam Shamim, Jonathan Alexander Gibson, Patrick Morrison,
    and Akond Rahman. 2022. \emph{Benefits, Challenges, and Research
    Topics: A Multi-vocal Literature Review of Kubernetes}.
  \item
    The Istio service mesh. Retrieved March 29, 2023 from \url{https://istio.io/latest/about/service-mesh/}
  \item
    Kubeflow Pipelines. Retrieved March 29, 2023 from \url{https://www.kubeflow.org/docs/components/pipelines/}
  \item
    Argo Workflows. Retrieved March 29, 2023 from \url{https://argoproj.github.io/argo-workflows/}
  \item
    Sherine Zhang and K. X. Zhang. 2019. PetFinder Challenge: Predicting
    Pet Adoption Speed.
  \item
    Terraform. Retrieved March 29, 2023 from
    \url{https://developer.hashicorp.com/terraform/docs}
  \item
    ArgoCD. Retrieved March 29, 2023 from
    \url{https://argo-cd.readthedocs.io/en/stable/}
  \end{enumerate}
\end{document}